\documentclass[letterpaper, 10 pt, conference]{ieeeconf}

\IEEEoverridecommandlockouts
\usepackage{amsmath}
\usepackage{amssymb}
\usepackage[geometry]{ifsym}
\usepackage{cite}
\usepackage{graphicx}
\newsavebox\mybox

\usepackage{ifsym}

\usepackage{caption}
\usepackage{subcaption}
\usepackage{array}
\usepackage[hidelinks]{hyperref}

\usepackage{enumerate}
\usepackage{algorithmic}
\usepackage[ruled,linesnumbered,noend]{algorithm2e}

\usepackage[normalem]{ulem}
\usepackage{color}
\usepackage{xcolor}

\newcommand{\snew}{\ensuremath{\boldsymbol{\zeta_i}}}
\newcommand{\Mnew}{\ensuremath{\Phi^{new}}}
\newcommand{\avec}{\ensuremath{\boldsymbol{\alpha}}}

\newcommand{\Bcurrent}{\ensuremath{B_i^{curr}}}

\title{\bf
Automated Task Updates of Temporal Logic Specifications for Heterogeneous Robots}
\author{Amy Fang and Hadas Kress-Gazit
\thanks{The authors are with the Sibley School of Mechanical and Aerospace Engineering, Cornell University, Ithaca, NY, 14853 USA. {\tt\small \{axf4,hadaskg\}@cornell.edu}. This work is supported by DARPA-PA-19-03-01 and the National Defense Science \& Engineering Graduate Fellowship (NDSEG) Program.
}}

\begin{document}
\maketitle
\thispagestyle{empty}
\pagestyle{empty}

\begin{abstract}
 Given a heterogeneous group of robots executing a complex task represented in Linear Temporal Logic, and a new set of tasks for the group, we define the task update problem and propose a framework for automatically updating individual robot tasks given their respective existing tasks and capabilities.  Our heuristic, token-based, conflict resolution task allocation algorithm generates a near-optimal assignment for the new task. We demonstrate the scalability of our approach through simulations of multi-robot tasks. 
\end{abstract}

% \begin{IEEEkeywords}
% Formal methods in robotics, multi-robot systems, task allocation, multi-robot teaming
% \end{IEEEkeywords}

\section{Introduction}
% include more citations
% add a multi-robot task allocation paragraph

Heterogeneous multi-robot systems consist of robots with different capabilities and are often created with a specific task in mind. However, if the task is changed during execution, especially if more requirements are added to the previous ones, there is a need for automated techniques that would allocate the task to the robots while maintaining the previous task and minimizing cost. For example, 
% \hkc{complete this} \amy{done} 
in humanitarian aid or disaster response situations, as new emergencies arise and timing is critical, automating the process for robots to interleave new tasks into their existing tasks without human input would 1) increase efficiency in the assignment process 2) ensure that the teams are responding quickly.

In this paper, we address the problem of automatically updating robot behaviors given tasks encoded in Linear Temporal Logic (LTL) over an abstraction of the robot motion and capabilities. We assume each sub-task can be accomplished by a single robot, and that all tasks are defined over controllable atomic propositions. We also assume there exist continuous controllers that can implement the abstract behaviors in a way that ensures collision avoidance between the robots and guarantees that the continuous behaviors implement the abstract ones~\cite{Kress-Gazit2018}. %   a discrete problem with abstracted robot capabilities and motion that can be implemented with continuous controllers.

There exists a rich literature in addressing automated multi-robot task allocation and coalition formation \cite{Gunn2015,Gerkey2004,Wang2017}. 
Researchers have developed architectures to model robot capabilities and interactions, such as social networks \cite{Klusch2002}. Game-theoretic models represent systems as stochastic games by using Markov Decision Processes (MDPs) or partially observable MDPs \cite{Wang2002}. While these methods can be shown to converge to a locally optimal policy, they do not scale for large numbers of agents, since the state space grows exponentially as more agents are introduced. The work in \cite{Qu2019} addresses scalability for the multi-agent MDP problem, but does so by assuming special dependence structures among robots.

% \hkc{what is missing? why do these approaches not work in our case and/or what are we borrowing from them?}

Dynamic coalition formation has been of particular interest, where autonomous robots cooperate to perform emerging time-varying tasks. Current methods include greedy approximate algorithms \cite{Service2011}, particle swarm optimization \cite{Xu2015}, and evolutionary algorithms \cite{Li2009}. Another method is to use market-based algorithms for robots to form teams based on the bids they make for the specified task \cite{Choi2010,Bing2018}. This requires a leader that acts as a mediator for the group. To maintain a flat hierarchy while still allocating tasks in a near-optimal manner, \cite{Xu2005} proposes a token-based framework, where tasks and resources are abstracted as tokens and passed locally among agents. Each agent decides whether to keep the token or to pass it to other agents. This decision requires information about how the token has been passed around to other agents. In this paper, we develop a token-based scheme for task allocation that is able to maintain near-optimal results without any token history information. 

To increase the complexity of possible tasks, in recent years there has been a growing interest in multi-robot planning for task specifications written in temporal logics,
% \hkc{explain why temporal logic is interesting} \amy{done}
which enables users to specify temporally extended tasks. In \cite{Mosca2019}, the authors use Time Window Temporal Logic to address multi-robot planning with synchronization requirements. The authors of \cite{Jones2019} encode Signal Temporal Logic specifications as mixed integer linear constraints to generate a plan for
a heterogeneous team. The work in \cite{Schillinger2018} presents an algorithm that allocates tasks to robots while simultaneously planning their actions. To avoid state-space explosion in their centralized planning, the authors sequentially link each robot model using switch transitions. In \cite{Schillinger2019}, the authors use reinforcement learning to synthesize plans over Markov decision processes under Linear Temporal Logic, and an auction-based algorithm assigns tasks to robots. All of these approaches assume that new tasks can only be directed towards unassigned robots. 

There also exists literature that addresses the problem of rescheduling in response to unexpected
disturbances. The authors of \cite{Wang2019} propose a bargaining game approach to generate a real-time scheduling scheme for an Internet of Things-enabled job shop. The work in \cite{Wong2006} presents an online hybrid contract-net negotiation protocol in response to unexpected disturbances in a shop. Agents place bids of their earliest finishing time, and a coordinator creates a new schedule accordingly.
While these papers address task reallocation due to unexpected environment changes, to our knowledge, no work has been done to tackle the problem of multi-robot task distribution within the context of temporal logics, where new tasks are introduced to robots that are performing existing tasks.

\textbf{Contributions}: In this paper, we propose a method for heterogeneous robots to respond to a new task given their capabilities and respective ongoing tasks while providing guarantees on  task feasibility. The contributions are as follows: 1) a mathematical formulation of the new task distribution problem, 2) a framework for robots to automatically update their behavior based on both the new task that is introduced and the progress within their current one, allowing them to perform both tasks, and 3) a heuristic, token-based task allocation algorithm to determine the final task allocation assignment for the new task while minimizing overall cost. 

\section{Preliminaries}

\subsection{Linear Temporal Logic}

Let $AP$ be a set of atomic propositions such that $\pi \in AP$ is a Boolean variable. We use these propositions to capture robot capabilities. For example, \textit{pick\_up} can correspond to the robot performing a pick up action.

\textbf{Syntax:} An LTL formula \cite{Baier2008} is defined recursively from atomic propositions $\pi \in AP$ using the following grammar:
\[
    \varphi ::=  \pi  \ |  \ \neg\varphi  \ | \ \varphi \vee \varphi \ |  \ \bigcirc \varphi  \ | \ \varphi \ \mathcal{U} \  \varphi
\]

\noindent where $\neg$ (``not") and $\vee$ (``or") are Boolean operators. $\bigcirc$ and $\mathcal{U}$ are the temporal operators ``next" and ``until", respectively.
Using these basic operators, we can construct the additional logical operators conjunction $\varphi \wedge \varphi$, implication $\varphi \Rightarrow \varphi$, and equivalence $\varphi \Leftrightarrow \varphi$, as well as the temporal operators eventually $\Diamond \varphi$ and always $\Box \varphi$.

\textbf{Semantics:} The semantics of an LTL formula $\varphi$ are defined over a trace $\sigma$, where $\sigma = \sigma_0\sigma_1\sigma_2...$ is an infinite sequence, and $\sigma_p$ represents the set of $AP$ that are True at position $p$. We denote that $\sigma$ satisfies LTL formula  $\varphi$ as $\sigma \models \varphi$. 

Intuitively, $\sigma \models \bigcirc \varphi$ if $\varphi$ is True at the next state in the trace. To satisfy $\varphi^1 \ \mathcal{U} \ \varphi^2$, $\varphi^1$ must stay True until $\varphi^2$ becomes True. The formula $\Box \varphi$ is satisfied if $\varphi$ is True at every position in $\sigma$, and $\sigma \models \Diamond \varphi$ if there exists a step in $\sigma$ where $\varphi$ is True. 
For a complete definition of the semantics of LTL, see \cite{Baier2008}. 

\subsection{B{\"u}chi Automata} \label{sec:buchi}

A nondeterministic B{\"u}chi automaton can be constructed from an LTL formula such that an infinite trace is accepted by the B{\"u}chi automaton if and only if it satisfies the LTL formula \cite{spot}. A B{\"u}chi automaton is defined as a tuple $B= (\Sigma, Z, z_0,  \delta, F)$, where $\Sigma$ is the alphabet of $B$, $Z$ is a finite set of states, $z_0 \in Z$ is the initial state, $\delta: Z \times \Sigma \rightarrow 2^Z$ is a transition function, and $F \subseteq Z$ is a set of accepting states. A run of a B{\"u}chi automaton  on an infinite word $w = w_1 w_2 w_3 ...$ is an infinite sequence of states $z = z_0 z_1 z_2...$ such that $\forall i, w_i\in\Sigma$ and $(z_{i-1}, w_i, z_{i}) \in \delta$. A run is accepting if and only if inf($z$) $\cap \ F \neq \emptyset$, where inf($z$) is defined as the set of states that are visited infinitely often in $z$ \cite{Baier2008}.

\textbf{B\"uchi intersection:}  Given two LTL formulas, $\varphi^1$ and $\varphi^2$ over $AP$, the intersection of their respective B\"uchi automata, $B^1$ and $B^2$, represents traces that satisfy both $\varphi^1$ and $\varphi^2$.

Let $B^1 = (\Sigma, Z_1, z_1^0, \delta_1, F_1),B^2 = (\Sigma, Z_2, z_2^0, \delta_2, F_2)$. Their intersection is defined as $B^1 \cap B^2 = (\Sigma, Z_1 \times Z_2 \times \{0,1,2\}, \{z_1^0, z_2^0, 0\}, \delta', Z_1 \times Z_2 \times \{2\})$. There is a transition on $a\in\Sigma$, $(\langle r,q,x \rangle, a, \langle r',q',x' \rangle) \in \delta'$ if and only if $(r, a, r') \in \delta_1$ and $(q, a, q') \in \delta_2$. The components $x, x' \in \{0,1,2\}$ are determined by $F_1$  and $F_2$, the accepting conditions of the B\"uchi automata. It is insufficient to simply define the accepting states of $B^1 \cap B^2$ as $F_1 \times F_2$ -- even if the accepting states from both automata appear individually infinitely often, they may appear together only finitely many times. Thus, $x, x' \in \{0,1,2\}$ ensures that accepting states from both $B^1$ and $B^2$ appear infinitely often together. For a detailed explanation on how to construct the intersection of two B\"uchi automata, see \cite{Clarke2000}.

\section{Problem Setup}
\subsection{Task Specification}\label{sec:task_spec}

A task is a set of LTL formulas $\Phi = \{\varphi^1, \varphi^2, ..., \varphi^m\}$ for which the following properties hold:

\begin{itemize}
    \item \textbf{\textit{Non-conflicting}}: There exists a $\sigma$ such that $\forall j, k$ $ \ \sigma \models \varphi^j \wedge \varphi^k$. 
    Intuitively, this means that the satisfaction of one sub-task must not violate any other sub-task.
    
    % \item \textbf{\textit{Independent}}: $\forall \varphi_j, \ \exists \ i$ such that $b_i \models \varphi_j$. That is, only one robot is required to satisfy a sub-task. \hkc{the name, formula and english explanation say different things. Here is an attempt at rewriting:}
 
     \item \textbf{\textit{Non-collaborative}}: Every $\varphi^j$ can be satisfied by a single robot. 
    % \item \textbf{\textit{Complete}}: a task specification is satisfied iff all sub-tasks are satisfied
\end{itemize}

\noindent \textit{Example.} The task ``pick up a box from room 2 and drop it off in room 3, pull the lever in room 3, and repeatedly scan and take a picture in room 1" can be encoded in LTL and decomposed into three sub-tasks: 

% \begin{equation}
% \neg drop\_off \ \mathcal{U} \ \big(\Diamond (room_1 \wedge pick\_up) \wedge \Diamond  (room_2 \wedge drop\_off)\big)
% \end{equation}

% F((r1 & p) & X(!d U (r2 & d)))
% \begin{equation}
% \begin{aligned}
% \Diamond &\big((room_1 \wedge pickup) \\ \wedge &\Diamond (\neg dropoff \ \mathcal{U} \ (room_2 \wedge dropoff))\big)
% \end{aligned}
% \end{equation}

%  !d U ((r1 & p) &F(r2 & d))
% \begin{equation}\label{new_mission}
\vspace{-0.75em}
\begin{align}\label{eq:m1}
    \varphi^1 = (&\neg \mathit{drop\_off} \ \mathcal{U} \
    (\mathit{room_2} \wedge \mathit{pick\_up)}) \\
    \wedge (&\neg \mathit{drop\_off} \ \mathcal{U} \  (\mathit{room_3} \wedge \mathit{drop\_off})) \notag
\end{align}

% \end{equation}
\vspace{-5mm}
\begin{align}
\varphi^2 &= \Diamond (\mathit{room_3} \wedge \mathit{pull\_lever}) \label{eq:m2} \\
% \end{equation}
\vspace{-5mm}
% \begin{equation}\label{eq:m3}
\varphi^3 &= \Box \Diamond (\mathit{room_1} \wedge \mathit{scan} \wedge \mathit{use\_camera}) \label{eq:m3}
\end{align}

% \noindent \textbf{Cost vector:} $\boldsymbol{c_i} = \langle c_{i}^1, c_{i}^2, ..., c_{i}^m \rangle$, where $c_{i}^j$ is the cost of $b_i$ that satisfies sub-task $j$ with minimum cost. Each $c_{i}^j$ is calculated using Equation \ref{eq_cost}.
% If a robot cannot perform a sub-task ($s_{i}^j = 0$), then $c_{i}^j = \infty$.

% \noindent \textbf{Assignment vector:} $\boldsymbol{a_i} = \langle a_{i}^1, a_{i}^2, ..., a_{i}^m \rangle$, where $a_i^j = 1$ if $A_i$ is assigned to perform sub-task $j$, and 0 otherwise.

\subsection{Robot Model}

% \begin{center}
%     \includegraphics[width=\columnwidth]{example-image-a}

%     \captionof{figure}{Capability Model}
%     \label{fig:robot}
% \end{center}

\begin{figure}
    \centering
    \newlength{\MyHeight}
    \settoheight{\MyHeight}{\includegraphics[scale = 0.2]{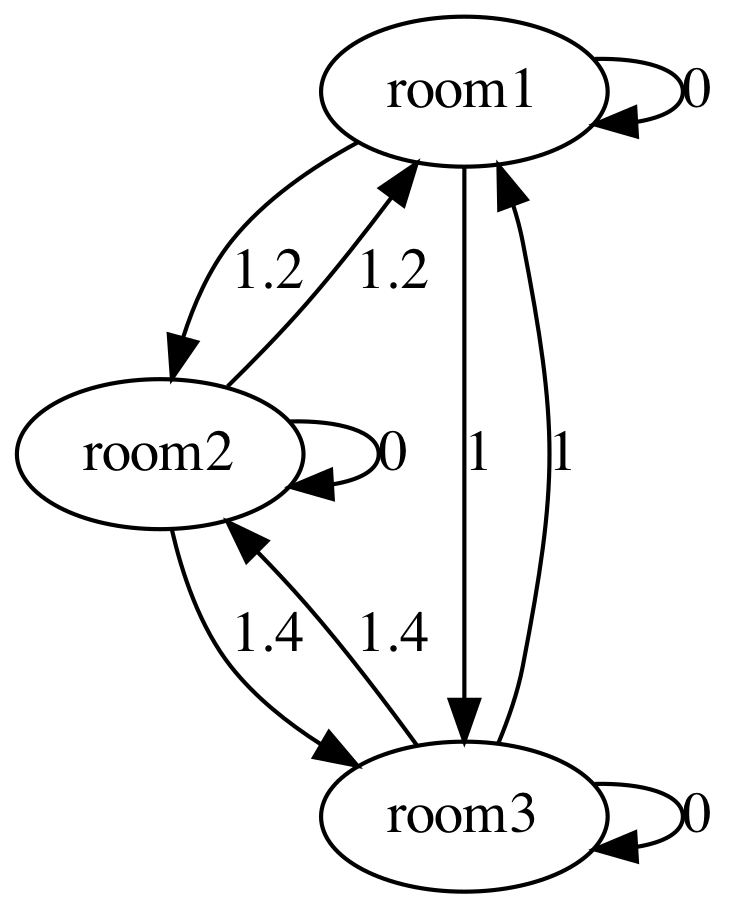}}
    
    \begin{subfigure}{0.47\columnwidth}
        \begin{minipage}[b][\MyHeight][c]{0.2\columnwidth}
            \includegraphics[scale=0.15]{figures/env_cap_v2.png}
        \end{minipage}
        \vspace{-5mm}
        \caption{}
        \label{fig:motion}
    \end{subfigure}%
    \begin{subfigure}{0.49\columnwidth}
        \begin{minipage}[b][\MyHeight][c]{0.2\textwidth}
            \includegraphics[scale=0.15]{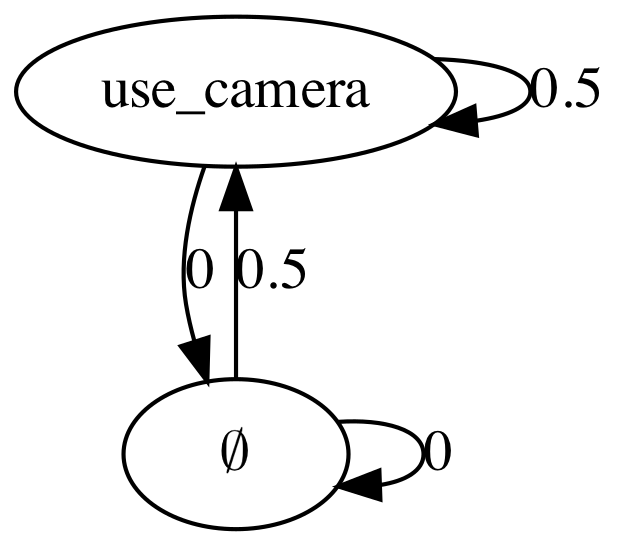}
        \end{minipage}
        \vspace{-5mm}
        \caption{}
        \label{fig:cam}
    \end{subfigure}
    
    \begin{subfigure}{\columnwidth}
        \centering
        \includegraphics[width=0.95\columnwidth]{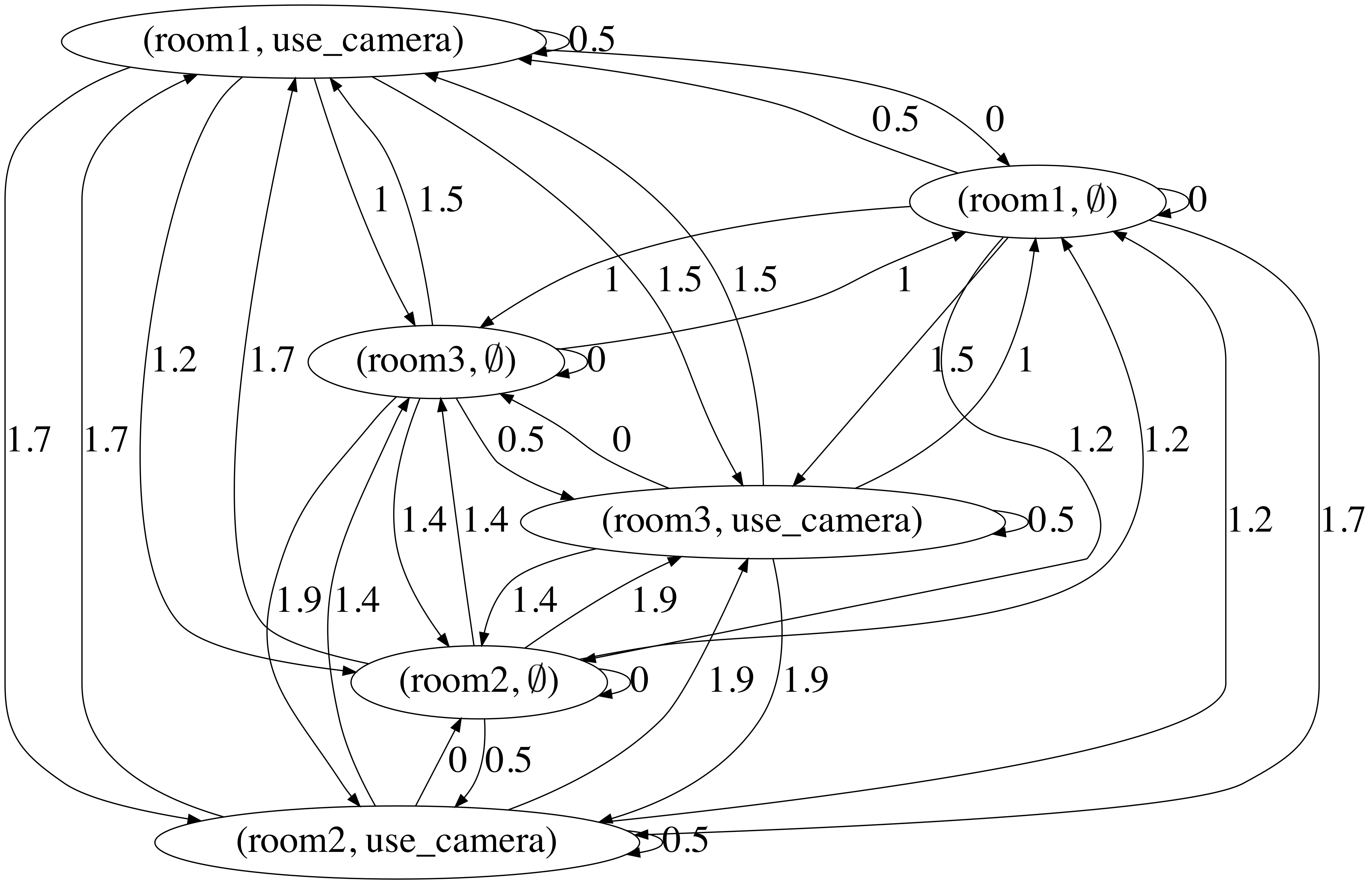}
        \caption{}
        \label{fig:cap_prod}
    \end{subfigure}
\vspace{-2mm}
\caption{Example of a motion model (\subref{fig:motion}), a capability (\subref{fig:cam}), and the robot model (\subref{fig:cap_prod}) %\amy{is this right now? I wasn't sure how to denote when camera = False. Right now I've made it $\emptyset$}
}
\label{fig:robot}
\vspace{-1em}
\end{figure}

Each robot in the group has a set of capabilities related to the required tasks. We define a \textit{capability} as a weighted transition system defined as a tuple $\mathcal{\lambda} = (AP, S, s_0, R, L, W)$, where

\begin{itemize}
    \item $AP$ is a set of atomic propositions
    \item $S$ is a finite set of states
    \item $s_0 \in S$ is the initial state
    \item $R \subseteq S \times S$ is a transition relation where for all $s \in S$ there exists $s' \in S$ such that $(s,s') \in R$
    \item $L: S \rightarrow 2^{AP}$ is the labeling function such that $L(s) \subseteq AP$ is the set of $AP$ that are true in state $s$
    \item $W: R \rightarrow \mathbb{R}_{\ge 0}$ is the cost function
\end{itemize}

Let there be a set of $v$ capabilities, $\Lambda = \{\lambda_1, \lambda_2, ... \lambda_v\}$ where $\lambda_k = (AP_k, S_k, s_{0,k}, R_k, L_k, W_k)$. We consider robots that are heterogeneous, where each robot has its own capability set $\Lambda_i \subseteq \Lambda$. 

We assume all robots are moving in a shared workspace that is partitioned into a set of regions. We describe the possible motion of a robot as a motion capability $\lambda_{\mathit{Mot}}$, where $AP$ is the set of region labels, and $(s, s') \in R$ if and only if a robot can directly move from the region labeled with $L(s)$ to the region labeled with $L(s')$.

A robot model $A_i$ is given by the product of its motion model and $u = |\Lambda_i|$ capabilities: $A_i = \lambda_{\mathit{Mot}} \times \lambda_1 \times ... \times \lambda_u$ such that $A_i=(AP, S, s_0, R, L, W)$, where 

\begin{itemize}
    \item $AP = \bigcup_{k=1}^u AP_k$
    \item $S = S_{\mathit{Mot}} \times S_1 \times S_2 \times ... \times S_u$ 
    \item $s_0 = (s_{0, \mathit{Mot}}, s_{0,1}, s_{0,2},..., s_{0,u})$
    \item $R \subseteq S \times S$ is a transition relation where for all $s = (s_{\mathit{Mot}}, s_1, s_2, ..., s_u)$ in $S$, there exists $s' = (s'_{\mathit{Mot}}, s'_1, s'_2, ..., s'_u)$ in $S$ such that $(s_{\mathit{Mot}}, s'_{\mathit{Mot}}) \in R_{\mathit{Mot}}, (s_1, s'_1) \in R_1, (s_2, s'_2) \in R_2, ..., (s_u, s'_u) \in R_u$.
    \item  $L$ is the labeling function such that $L(s) = (L_{\mathit{Mot}}(s_{\mathit{Mot}}), L_1(s_1), L_2(s_2), ... , L_k(s_k))$  
    \item $W: R \rightarrow \mathbb{R}_{\ge 0}$ is the cost function
\end{itemize}

\noindent Let $s = (s_{\mathit{Mot}}, s_1, s_2, ..., s_u)$ and $s' = (s'_{\mathit{Mot}}, s'_1, s'_2, ..., s'_u)$ be two states in $A_i$ for which the transition $(s, s')$ exists. Then, the cost $W((s, s'))$ is the sum over $W((s_{\mathit{Mot}}, s'_{\mathit{Mot}}))$ and all $W((s_k, s'_k))$.
See Fig. \ref{fig:robot} for a simple example. 
In this motion model, the cost for transition $(room_1, room_2)$ is $W_{\mathit{Mot}}((room_1, room_2)) = 1.2$, and the cost for the transition $(\emptyset, use\_camera)$ in the capability is $W_1((\emptyset, use\_camera)) = 0.5$. Then, 
in the robot model, the cost of the transition between states $(room_1, \emptyset)$ and $(room_2, use\_camera)$ is 1.7. 

% The alphabet $\Sigma = 2^{AP}$ is consistent between the robot model and the missions, which are written in LTL. For example, for a task that requires picking up a box, the mission will be encoded with the proposition "use arm." (Future research will allow for different alphabets, where the higher level mission specifications can automatically be translated into the capability alphabet.)
\subsection{Robot Behavior}\label{sec:calcs}
Given a robot model $A_i$ and a desired behavior $\varphi$ captured using $B^\varphi$, we can synthesize a behavior for the robot such that it satisfies $\varphi$ by choosing an accepting trace in $\mathcal{G} = A_i \times B^\varphi$ \cite{Baier2008}. %\hkc{define this product automaton, including the accepting condition and the weights} \amy{done} \hkc{add details}
%\amy{added: }
Let $A_i = (AP^i,S,s_0,R,L,W)$ and $B^\varphi= (\Sigma, Z, z_0,  \delta, F)$ such that $\Sigma = 2^{AP^\varphi}$ is created from $AP^\varphi$, the propositions in $\varphi$. 
Note that it is not necessary for $AP^i$ to be equivalent to $AP^\varphi$. $AP^i \nsubseteq AP^\varphi$ when a robot has additional capabilities that are not required for task $\varphi$. Similarly, $AP^\varphi \nsubseteq AP^i$  when a robot does not have all the necessary capabilities required for task $\varphi$. 

%For the following, we define the projection operator: given two sets of symbols $a,b$  $a \Big\rvert_{b} = \sigma \cap AP^i$ that returns the set of propositions in $\sigma$ that also appear in $AP^i$. % We check for this condition by evaluating $\Sigma$ at the labeling function $L$, $\Sigma \Big\rvert_{L(s)}$.

% \amy{to do: preface that APs don't have to be the same, explain projection operator, explain what happens if sigma has things that are not in AP}

The product $\mathcal{G} = (\Sigma, AP^i, Q, q_0, L^\mathcal{G}, \Delta, W^\mathcal{G}, F^\mathcal{G})$, where

\begin{itemize}
    % \item $\Sigma^\mathcal{G} = \Sigma \cup 2^{AP}$ is a set of input alphabets 
    \item $Q = S \times Z$ is a finite set of states
    \item $q_0 = (s_0, z_0)$ is the initial state
    \item $L^\mathcal{G}$ is the labeling function such that $L^\mathcal{G}((s,z)) = L(s) \cap AP^\varphi$ %is the set $\Sigma \Big\rvert_{L(s)}$ 
    % for state $q = (s,z)$, $L^\mathcal{G}(q) = L(s)$
    \item $\Delta$ is the transition function, where a transition exists from $(s,z)$ to $(s',z')$ if and only if $(s,s') \in R$, and $\exists \ \sigma \in \Sigma$ such that $\sigma = L^\mathcal{G}(s')$ and $z' \in \delta(z, \sigma)$
    
    \item $W^\mathcal{G}: R \rightarrow \mathbb{R}_{\ge 0} $ is the cost function such that for $q = (s, z)$ and $q' = (s',z')$, $W^\mathcal{G}((q, q'))=W((s, s'))$
    \item $F^\mathcal{G} = S \times F$ is a set of accepting states
\end{itemize}

\noindent Let path $q = q_0 \rightarrow q_1 \rightarrow q_2 \rightarrow ... \rightarrow q_\ell \rightarrow q_{\ell+1} ... $ be an infinite sequence in $\mathcal{G}$ that visits the states in $F^\mathcal{G}$ infinitely often. The path is composed of a prefix -- a finite trace -- and a suffix -- a cycle that repeats. A behavior $b_i$ of a robot is defined as the labels produced by $q$: $b_i = L^\mathcal{G}(q_0 )L^\mathcal{G}(q_1)L^\mathcal{G}(q_2)...L^\mathcal{G}(q_\ell)L^\mathcal{G}(q_{\ell+1}) ...$ . Given a prefix of length $\ell$, %We assume that the behavior up to index $\ell$ is the prefix, and the remaining portion of the sequence is the suffix.   \hkc{why l? you need to write this as an assumption - we assume a behavior is composed of two parts, the prefix and suffix such that...} \amy{done?} The total 
we define the cost of $b_i$ as: %is defined up to $q_\ell$:
% \vspace{-0.5em}
\begin{equation}
    c_i(b_i) = \sum_{r=0}^{\ell-1} W^\mathcal{G}((q_{r}, q_{r+1}))
    \label{eq_cost}
\end{equation}
\vspace{-0.5em}
\begin{figure}[t]
\vspace{0.5em}
     \centering
     \includegraphics[width=0.6\columnwidth]{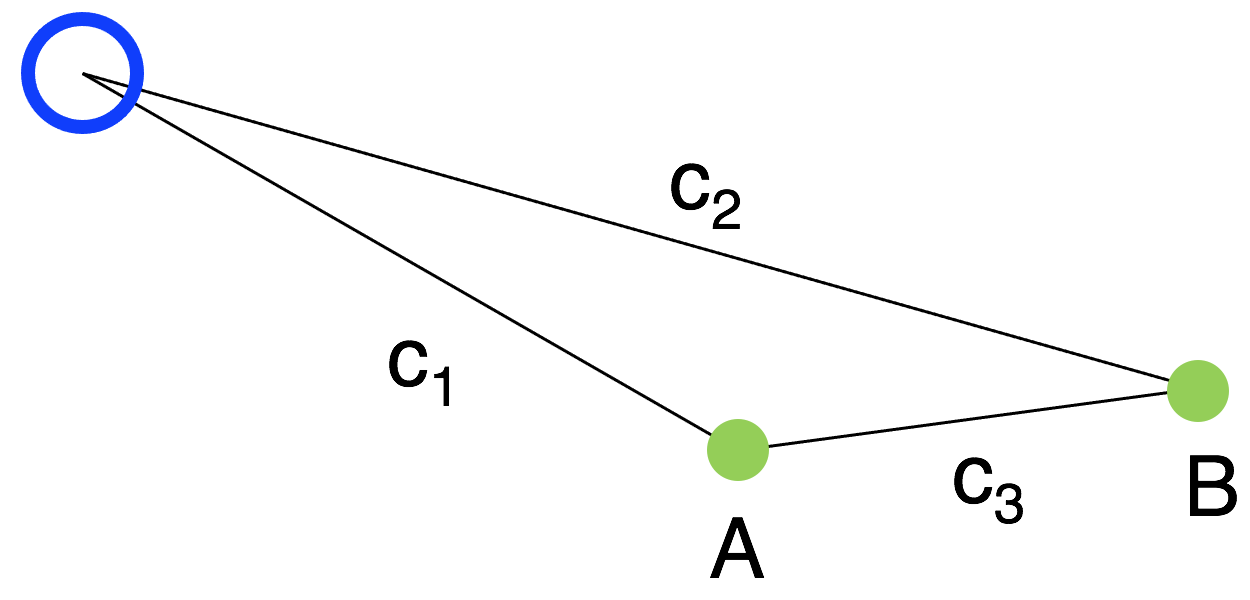}
    \caption{Example of non-additive cost. The robot, in blue, is tasked to go to points A and B. Performing the tasks at the same time costs less than doing them separately: $c_1 + c_3 < c_1 + c_2$.}
    \label{fig:cost}
\vspace{-1.1em}
\end{figure}

We allow the cost for each sub-task to be non-additive. That is, the sum of the costs of the behaviors for satisfying two individual sub-tasks may be more than the cost of the behavior for satisfying both. For a simple illustration, see Fig. \ref{fig:cost}.

We define $P$ to be a partition of \Mnew \ such that the following properties hold for $p_x \in P$:
% \vspace{-0.5em}
\begin{align}\label{eq:partition}
    p_x \cap p_y &= \emptyset, \quad \forall x \neq y \\ 
    \nonumber
    \bigcup_{x} p_x &= P
\end{align}
\vspace{-0.5em}

Given a set of new sub-tasks $\Mnew = \{\varphi^1, \varphi^2, ..., \varphi^m\}$, we define for each robot $A_i$ the corresponding cost and satisfiability structures: 

\noindent \textbf{Cost $\Gamma_i$} is a function that maps $\varphi_i^{curr} \bigwedge_{j \in p_i} \Mnew[j]$ to its corresponding cost, $c_i(b_i^{new})$, where $p_i \in P$ is the set of sub-tasks assigned to $A_i$. 

\noindent \textbf{Satisfiability $\snew$} denotes which sub-tasks $A_i$ is able to perform: 
\vspace{-0.5em}
\begin{equation*}
\zeta_{i}[j] = \begin{cases}
1 &\text{$\exists \ b_i$ such that $b_i \models \varphi^j$}\\
0 &\text{otherwise}
\end{cases}
\end{equation*}

\section{Problem Statement}

Let there be $n$ heterogeneous robots $A = \{A_1, A_2, ..., A_n\}$. Each robot $A_i$ has behavior $b_i^{curr}$ that satisfies its existing task specification $\varphi_i^{curr}$.

Given a new task \Mnew$= \{\varphi^1, \varphi^2, ..., \varphi^m\}$ that is introduced during the robots' execution of their current tasks, find a partition $P$, as defined in Eq. \ref{eq:partition}, to assign robots to new sub-tasks so that $b_i^{new} \models \varphi_i^{curr} \bigwedge_{j \in p_i} \varphi^j$, subject to the following optimization criteria:
\vspace{-0.5em}
\begin{equation}
    \mathop{\arg \min}\limits_{p_i \in P} \sum_{i=1}^{n} c_i(b_i^{new}),
\end{equation}

% \hkc{separate assumptions from what is given}. \amy{done}
We make the following assumptions about the system: collision avoidance is taken care of by low-level controllers; 
% there is no interference between robot behaviors, so that robots do not need to consider the behaviors of other robots when executing their own; 
the sub-tasks are nonreactive, meaning that the robot behavior does not depend on external events; each sub-task can be satisfied by a single robot and does not require robot collaboration, as outlined in \ref{sec:task_spec}.

\subsection{Example} \label{sec:example}

Consider a 2D environment with five rooms containing three robots.
% \hkc{what is "usecamera"? what is "armidle"? it would be good to mention $AP$ before giving details regarding the robots} \amy{done}
The set of all capabilities is $\Lambda = \{ \lambda_{\mathit{Mot}}, \lambda_{arm}, \lambda_{scan}, \lambda_{camera} \}$. $AP_{arm}$ is an abstraction of a physical robot manipulator that is capable of grasping objects, such as boxes and levers.  $AP_{scan}$ represents a robot's ability to scan barcodes. Similarly, we abstract a robot's camera as $AP_{camera}$, which denotes whether or not the robot is taking a picture.

Each robot has the following capabilities and tasks:

\begin{itemize}
    \item Robot 1: ``Scan in room 4, then scan in room 1"
    \vspace{-0.5em}
        \begin{multline}\label{eq:curr1}
            \varphi_1^{curr} = (\neg(room_1 \wedge scan) \ \mathcal{U} \ (room_4 \wedge scan)) \\
            \wedge \Diamond(room_1 \wedge scan) 
        \end{multline}
        \vspace{-2em}
        \begin{align*}
            \Lambda_1 = \{\lambda_{\mathit{Mot}}, \lambda_{arm}, \lambda_{scan}\}, s_0 = (room_2, \emptyset, \emptyset)
        \end{align*}
        
    \item Robot 2: ``Repeatedly travel between rooms 2 and 5 and scan in those rooms"
    \vspace{-0.5em}
        \begin{multline}\label{eq:curr2}
            \varphi_2^{curr} = \Box \Diamond (room_2 \wedge scan) \\
            \wedge \Box \Diamond (room_5 \wedge scan)
        \end{multline}
        \vspace{-2em}
        \begin{align*}
            \Lambda_2 = \{\lambda_{\mathit{Mot}}, \lambda_{scan}, \lambda_{camera}\}, s_0 = (room_1, \emptyset, \emptyset)
        \end{align*}
        
    \item Robot 3: ``Take a picture in room 1 and pick up a box in room 4, in any order"
    \vspace{-0.5em}
        \begin{multline}\label{eq:curr3}
            \varphi_3^{curr} = \Diamond(room_1 \wedge use\_camera)  \\
            \wedge \Diamond (room_4 \wedge pickup)
        \end{multline}
        \vspace{-2em}
         \begin{align*}
            \Lambda_3 = \{\lambda_{\mathit{Mot}}, \lambda_{arm}, \lambda_{camera}\}, s_0 = (room_2, \emptyset, \emptyset)
        \end{align*}
\end{itemize}

The new task \Mnew, provided in Eq. \ref{eq:m1} - \ref{eq:m3}, is introduced while the robots are executing these tasks.

\section{Approach}
%\hkc{need a few sentences summarizing the approach and this section} \amy{done?}

The approach is as follows: each robot determines how much of the current task it has already accomplished. There are two reasons for this: 1) so that the robot does not repeat completed portions of the task (thus reducing cost), and 2) so that if the new task conflicts only with completed portions of the current task, the robot does not deem the new task as impossible to achieve. The robot then synthesizes the corresponding behavior for the new sub-tasks based on its capabilities and the remaining current task. It calculates the cost of performing different feasible combinations of sub-tasks (Sec. \ref{sec:synthesis}). To determine the assignment of tasks that minimizes the overall cost for the robots, we develop a token-based, conflict resolution task allocation algorithm. Robots pass around an assignment token and assign themselves to tasks based on the cost of the corresponding behavior (Sec. \ref{sec:allocation}).
\vspace{-0.15em}

\subsection{Synthesis of Robot Behavior}\label{sec:synthesis}
\vspace{-1em}
\begin{algorithm}[h]
    \SetKwInOut{Input}{Input}
    \SetKwInOut{Output}{Output}
    \SetKwProg{Initialization}{Initialization}{}{}
    \Input{$A_i$, $z_{i}$, \Bcurrent, $\varphi^k$ }
    \Output{$b_i^{new}, c_i(b_i^{new})$}

    $B^{k} := \textsc{ltl2buchi} (\varphi^k)$ \\
    $B_i^{curr} := \textsc{find\_reachable\_buchi}(z_{i}, B_i^{curr})$ \\
    $B_{i}^k := \textsc{create\_buchi\_intersect}(B^{k}, B_{i}^{curr})$ \\

    % \amy{***}
    % \If{$A_i \times B_i^k \neq \emptyset$}{
        $\mathcal{G} = A_i \times B_{i}^k$ \\
        \tcp{Let $F^\mathcal{G} :=$ be the set of accepting states in $\mathcal{G}$}  
        \tcp{Let $q_i :=$ initial state of $\mathcal{G}$}
        $b_i^{new}, c_i(b_i^{new}) := \textsc{dijkstra}(\mathcal{G}, q_i, F^\mathcal{G}) $ \\
        \If{$b_i^{new} = \emptyset$}{
            $c_i(b_i^{new}) := \infty$ }
    % }
    % \Else{$b_i^{new} := \emptyset$}
\caption{Synthesize Behavior}
\label{algo:synthesis}
\end{algorithm}
\vspace{-0.5em}

Given a new task, each robot runs Alg.~\ref{algo:synthesis} to automatically synthesize a new behavior. We transform the LTL formula $\varphi^k$ into B\"uchi automaton $B^k$ using Spot~\cite{spot} (line 1). 

To synthesize a behavior for a robot that would cause it to perform both its current task and $\varphi^k$, we first determine what the robot needs to do to complete its current task. To do so, we  calculate the reachable portion of $B_i^{curr}$ %We do this because we need to find a new behavior only for what the robot has yet to accomplish for its current task, rather than the entire task specification. In addition, if $\varphi_i^{curr} \wedge \varphi^k$ is a contradictory formula, it may still be possible for the robot to satisfy both tasks. The robot may have already performed enough of $\varphi_i^{curr}$ prior to receiving $\varphi^k$ such that there are no longer any conflicts between the two tasks. As such, when taking the intersection with $B^k$, we consider only the reachable portion of $B_i^{curr}$ 
from the state the robot is in when the new task is introduced, denoted as $z_{i}$. To generate the reachable portion of $B_i^{curr}$, the function \textsc{find\_reachable\_buchi} calculates the forward reachable set \cite{Alur2000} 
% \hkc{i do not think this is a good citation - isn't it for continuous systems?} 
and removes any non-reachable states. 

The function \textsc{create\_buchi\_intersect} finds the intersection of $B_i^{curr}$, the current task remaining, and $B^{k}$ representing the new task (line 3). The alphabets $\Sigma_i^{curr}=2^{AP_i^{curr}}, \Sigma^k=2^{AP^k}$ of the respective B\"uchi automata might not be equivalent, since the task specifications $\varphi$ may require different capabilities and thus be defined over different $AP$s. 

Borrowing from the definition in Sec. \ref{sec:buchi}, the B\"uchi intersection $B_i^k$ has the alphabet $\Sigma_i^k = 2^{AP_i^{curr}\cup AP^k}$. % To define the transition function we consider three sets: $AP_1 = AP_i^{curr} \cap  AP^k$ the set of propositions in both $\Sigma_i^{curr}$ and $\Sigma^k$, $AP_2 = AP_i^{curr} \setminus  AP_1$, the set of propositions used in $\Sigma_i^{curr}$ but not in $\Sigma^k$, and $AP_3 = AP^k \setminus  AP_1$, the set of propositions used in $\Sigma^k$ but not in $\Sigma_i^{curr}$ 
Given $\sigma \in \Sigma_i^k$, a transition $(\langle r,q,x \rangle, \sigma, \langle r',q',x' \rangle) \in \delta_i^k$ if and only if $(r,\sigma \cap AP_i^{curr} ,r') \in \delta_i^{curr}$ and $(q,\sigma \cap AP^k,q') \in \delta^k$. All other elements in the tuple $B_i^k$ remain the same as defined in Sec. \ref{sec:buchi}.  

% \begin{itemize}
%     \item $\sigma_1 = 2^{AP_i^{curr} \setminus  AP^k}$, $(\langle r,q,x \rangle, \sigma_1, \langle r',q,x' \rangle) \in \delta_i^k$ iff $(r,\sigma_1,r') \in \delta_i^{curr}$
%     \item $\sigma_2 = 2^{AP^k \setminus  AP_i^{curr}}$, $(\langle r,q,x \rangle, \sigma_2, \langle r,q',x' \rangle) \in \delta_i^k$ iff $(q,\sigma_2,q') \in \delta^k$
%     \item $\sigma_3 = 2^{AP_i^{curr} \cap AP^k}$,
%     $(\langle r,q,x \rangle, \sigma_3, \langle r',q',x' \rangle) \in \delta_i^k$ iff $(r,\sigma_1 \cup \sigma_3,r') \in \delta_i^{curr}$ and $(q,\sigma_2 \cup \sigma_3,q') \in \delta^k$
% \end{itemize}

% \noindent All other elements in the tuple $B_i^k$ remain the same as defined in Sec. \ref{sec:buchi}.

% based on the definition in Sec. \ref{sec:buchi}. %\hkc{this needs a lot more details and explanations. Right now it reads as a "by the way...". Explain how you create the reachable buchi, what it means and why you are doing it (and hence how it can allow contradicting specifications)} \amy{done?}

In line 5, 
% \hkc{it is best to put labels on the lines and then ref them so there are less mistakes like this. not critical now, but for the future}
the robot calculates the minimum cost behavior $b_i^{new}$ by using Dijkstra's shortest path algorithm to find the minimum cost path through $A_i \times B_i^k$ to an accepting state and an accepting cycle. $b_i^{new} = \emptyset$ when the robot is unable to perform $\varphi^k$. This occurs either if the robot does not have the capabilities to satisfy the new task, or if the current remaining task and the new task conflict with each other.

Given $b_i^{new}$, each robot calculates its satisfiability \snew \ and cost $\Gamma_i$, as outlined in Alg. \ref{algo:vectors}. In lines 2-4, the robot determines if it can perform both its current task and the $j^\text{th}$ new sub-task. If it can, we set $\snew[j] = 1$ and include the corresponding cost in  $\Gamma_i$.

After calculating \snew, the robot synthesizes the behavior for each combination of sub-tasks it can do (lines 7-11). It does this because the cost for each sub-task is non-additive (as explained in Sec. \ref{sec:calcs}). The combinations of sub-tasks are determined based on \snew. 
\vspace{-0.5em}
\begin{algorithm}[h] 
    \SetKwInOut{Input}{Input}
    \SetKwInOut{Output}{Output}
    \SetKwProg{Initialization}{Initialization}{}{}
    \Input{$A_i$, $z_{i}$,  \Bcurrent, $\Mnew$ }
    \Output{$\snew, \Gamma_i$}
    $\snew := \mathbf{0}, \Gamma_i := \emptyset$ \\
    \tcp{for individual sub-tasks}
    \For{$j \in |\Mnew|$}{
        $b_i^{new}, c_i(b_i^{new}) := \textsc{synthesize\_behavior}(A_i, z_{i}, \Bcurrent, \Mnew[j])$ \\
        $\Gamma_i[j] := c_i(b_i^{new})$\\
        \If{$b_i^{new} \neq \emptyset$}{
            $\snew [j] := 1$ \\
        }
    }

    % $S := \textsc{satisfiable\_missions}(\snew)$ \\
    \mbox{\tcp{for combinations of sub-tasks}}
    $Y_i = \{j \ | \ \zeta_i[j] = 1\}$ \\ 
    \For{$k \in 2^{Y_i}$}{
        \If{$|k| > 1$}{
            $\varphi^k :=  \bigwedge_{\ell} \Mnew[k_\ell]$ \\
            $b_i^{new}, c_i(b_i^{new}) := \textsc{synthesize\_behavior}(A_i, z_{i}, \Bcurrent, \varphi^k)$ \\
             $\Gamma_i[k] := c_i(b_i^{new})$
            
        }
    }

    % \hspace*{\algorithmicindent} \For{$A_i \in A$}
\caption{Compute Satisfiability and Cost} 
\label{algo:vectors}

\end{algorithm}

\vspace{-1.5em}
\subsection{Task Allocation}\label{sec:allocation}

We introduce a token-based heuristic algorithm to determine a near-optimal allocation for the new task, as shown in Alg. \ref{algo:team}. The token is \avec, the global assignment vector of length $m$ where $\alpha_j $ corresponds to the robot that has been assigned to $\Mnew[j]$. $\avec$ is initialized to be a zero vector.

Each robot $A_i$ assigns itself to the sub-tasks that it can perform and have not yet been assigned to any other robot (lines 3-5). The algorithm includes a conflict resolution scheme when the sub-tasks that $A_i$ can perform have already been assigned (lines 6-16). 
In this case, for each robot $A_k$ with conflicts, the algorithm looks at the overlap between the tasks already assigned to $A_k$ and the tasks $A_i$ can do, which is provided by its satisfiability vector $\boldsymbol{\zeta_i}$ (lines 10-15). The function \textsc{update\_assignment} iterates through every combination of overlapping assignments for the $A_i$ and $A_k$, finds the one with the minimum cost, and updates \avec.

In this algorithm, at each iteration of the conflict resolution, a robot only compares possible conflicting assignments with one other robot. Although we cannot guarantee optimality of the final task allocation assignment, it significantly reduces the computation time. Our algorithm has complexity $\mathcal{O}(2^{m}n)$, compared to the optimal algorithm, which checks every ${n \choose m}$ combinations and has complexity $\mathcal{O}(m^n)$. In addition, because the token is passed to every robot, we can guarantee that the algorithm will find an assignment for every sub-task if one exists.
\vspace{-0.5em}

\begin{algorithm} 
    
    \SetKwInOut{Input}{Input}
    \SetKwInOut{Output}{Output}
    \SetKwProg{Initialization}{Initialization}{}{}
    \Input{$\boldsymbol{\zeta_1}, \boldsymbol{\zeta_2}, ..., \boldsymbol{\zeta_n}, \Gamma_1, \Gamma_2, ..., \Gamma_n $, $m := |\Mnew|$ }
    \Output{\avec}
     $\avec := \mathbf{0}$ \\
   
    \For{$i \in \{1,...,n \}$}{
        \For{$j \in \{1,...,m \}$}{
          \If{$\snew[j] = 1$ and $\avec[j] = 0$}{ 
            $\avec[j] = i$ 
            }
        % \tcp{If $A_i$ can satisfy $M^{new}[j]$ and a robot is assigned to it }
        }
        \tcp{Conflict resolution}
        $compared = \emptyset$ \\
        $satis_i := \{p \ | \ \zeta_i[p] = 1 \}$ \\
        \For{$j \in \{1,...,m \}$}{
        \If{$\snew[j] = 1$ and $\avec[j] \neq i$ }{
            $k =\avec[j] $ \\
            \If{$k \not\in compared$}{
                $assigned_i :=  \{p \ | \ \avec[p] = i \}$ \\
                $assigned_k :=  \{p \ | \ \avec[p] = k \}$ \\
            
                $conflicts := satis_i \cap assigned_k$ \\
                
                $\avec := \textsc{update\_assignment}(assign_i, assign_k,$  \hskip \algorithmicindent  $satis_i, conflicts, \Gamma_i, \Gamma_k)$
                
                $compared = compared \cup \{k\}$
                }
            }
        }
    }    
    % \hspace*{\algorithmicindent} \For{$A_i \in A$}
\caption{Task allocation for $\Mnew$} 
\label{algo:team}
\end{algorithm}
\vspace{-1.5em}
\section{Results and Evaluation}
% \vspace{-0.5em}
% \hkc{a few sentences about what you are going to show} \amy{done}
We demonstrate the effectiveness of our synthesis framework by showing the changes to the robot behaviors for the example in Sec. \ref{sec:example}. Furthermore, we compare the results of our token-based algorithm to the optimal assignment for different team and task sizes. %by running randomized simulations of different robots and new tasks, and compare the results against the optimal  assignments. 
% \vspace{-0.5em}
\subsection{Simulation of Robot Behavior}
\begin{figure}[h!]
\vspace{0.5em}
    \centering
    \includegraphics[width=0.85\columnwidth]{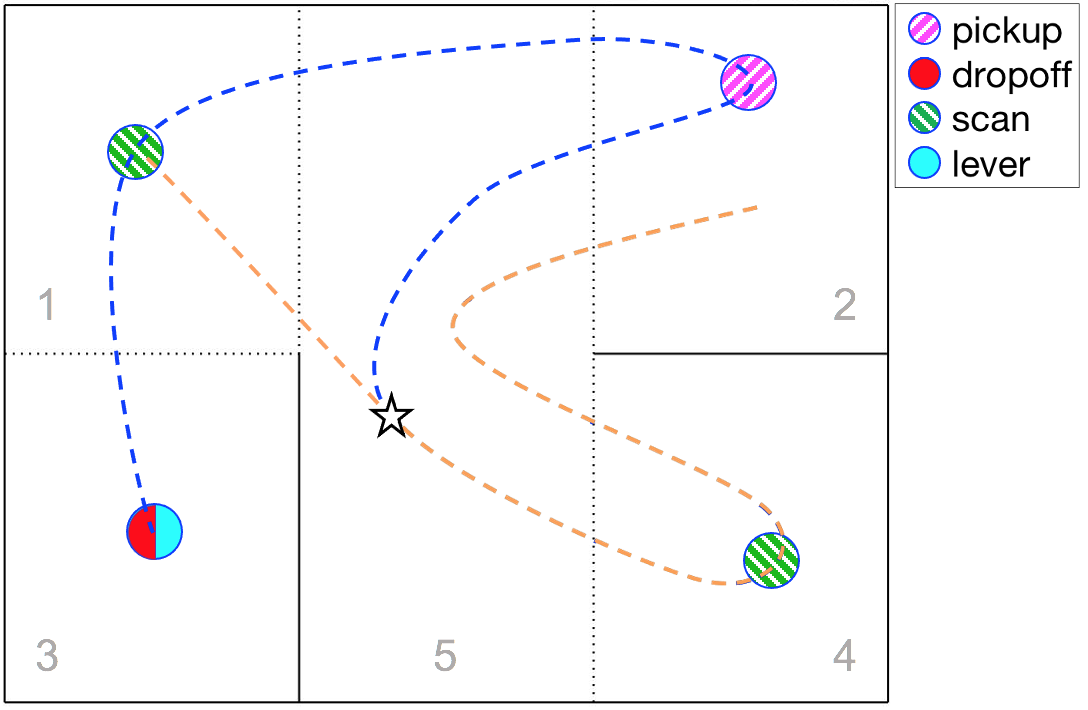}
    \caption{Updated behavior for Robot 1. The robot starts in room 2, and its original trajectory for the current task is drawn in orange. It receives the new task at the star and updates its behavior based on the sub-tasks it assigned itself. The new behavior is shown in blue. The colored circles indicate the action the robot takes.}
    \label{fig:ag3}
    \vspace{-1em}
\end{figure}

% \vspace{-0.5em}
For the example in \ref{sec:example}, the final task allocation assignment is $\avec = [1, 1, 2]$, meaning that Robot 1 is tasked to complete the first two sub-tasks (Eq. \ref{eq:m1}, \ref{eq:m2}), and Robot 2 is tasked to complete the third sub-task (Eq. \ref{eq:m3}). Fig. \ref{fig:ag3} shows the updated behavior of Robot 1 after being assigned, mid-execution, the new sub-tasks. For reference, its current task is to scan first in room 4 and then scan in room 1 (Eq. \ref{eq:curr1}). Observe that the robot interleaves the current and new tasks rather than performing them sequentially - before completing its current task by scanning in room 1, it performs part of the new task by picking up a box in room 2.

% \vspace{-1em}
\subsection{Task Allocation Performance}
\begin{figure}
\vspace{0.1em}
     \centering
     \begin{subfigure}[t]{\columnwidth}
         \centering
         \includegraphics[width=0.81\textwidth]{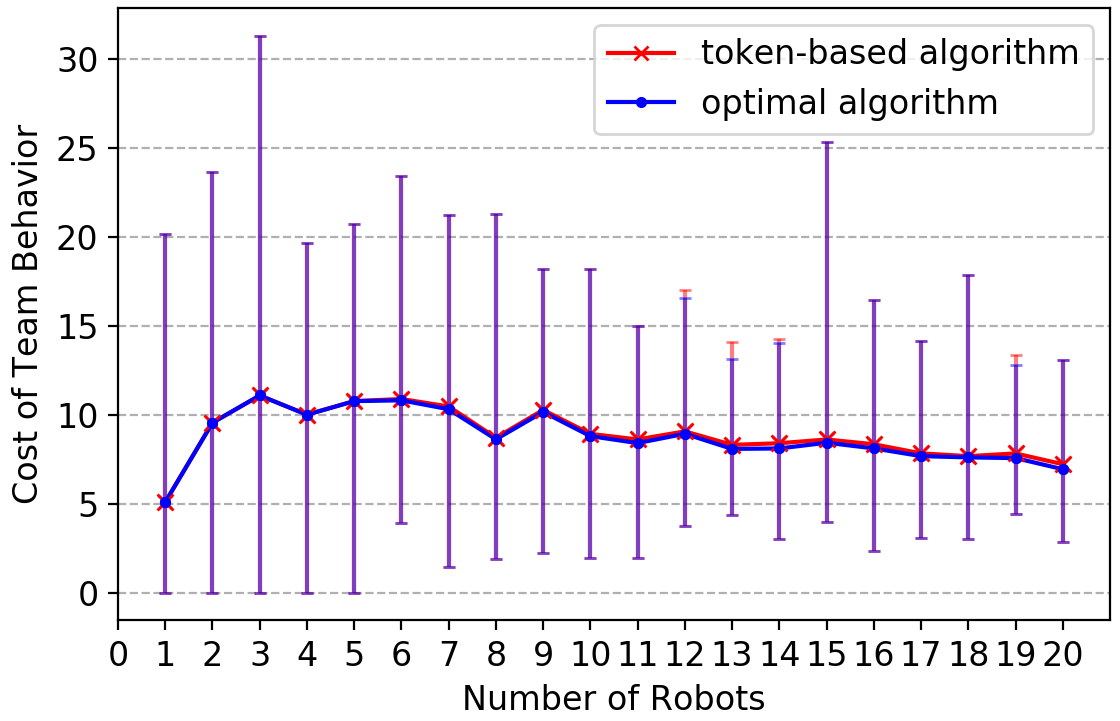}
         \caption{}
         \label{fig:time_robots}
     \end{subfigure}
    \centering
     \begin{subfigure}[t]{\columnwidth} 
         \centering
         \includegraphics[width=0.82\textwidth]{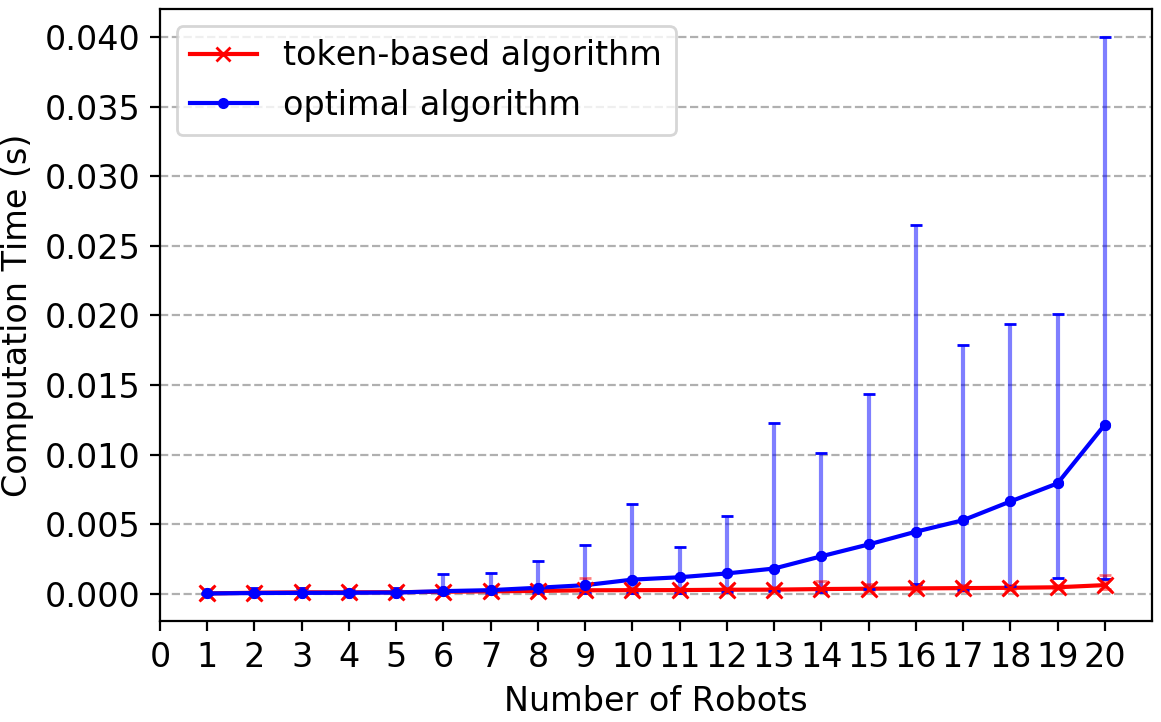}
         \caption{}
         \label{fig:cost_robots}
     \end{subfigure}
    \vspace{-0.5em}
    \caption{Comparison of overall cost (\subref{fig:cost_robots}) and computation time (\subref{fig:time_robots}) between our algorithm and the optimal algorithm when varying the number of robots. The error bars represent the min/max values of the simulations.} 
    \label{fig:vary_robots}
    
\end{figure}

\begin{figure}
\vspace{-1.75em}
     \centering
     \begin{subfigure}[t]{\columnwidth}
         \centering
         \includegraphics[width=0.8\textwidth]{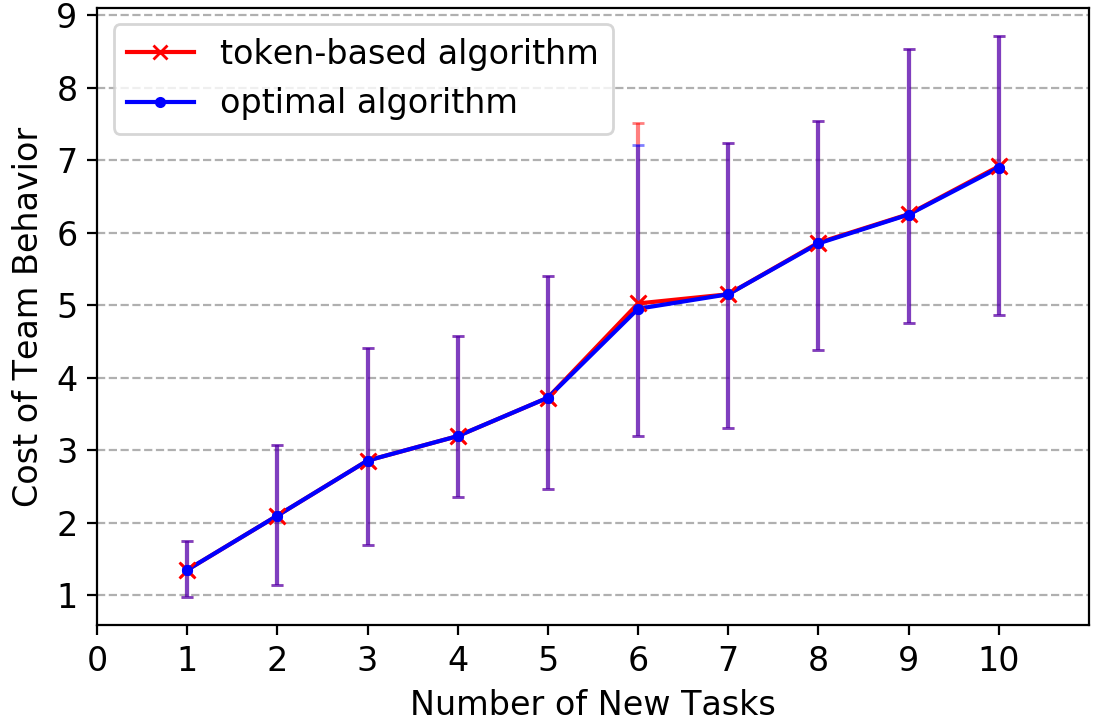}
         \caption{}
         \label{fig:cost_robots}
     \end{subfigure}
    \centering
     \begin{subfigure}[t]{\columnwidth} 
         \centering
         \includegraphics[width=0.82\textwidth]{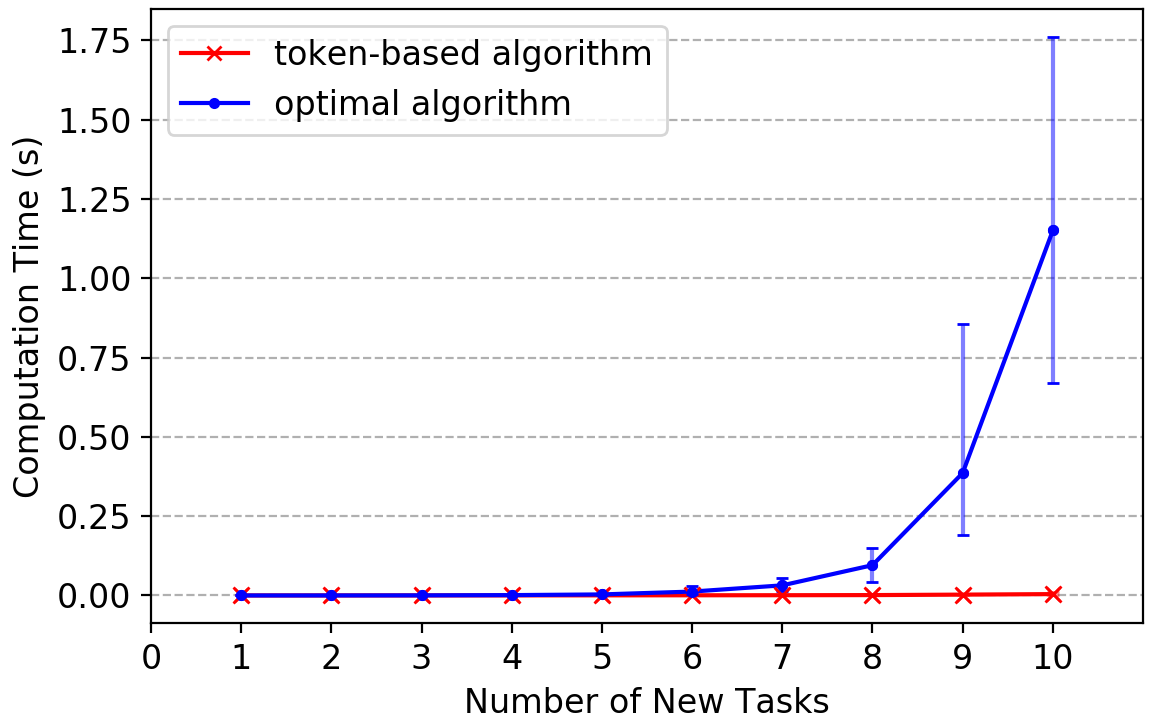}
         \caption{}
         \label{fig:time_robots}
     \end{subfigure}
    \vspace{-0.5em}
    \caption{Comparison of computation time (\subref{fig:cost_robots}) and cost (\subref{fig:time_robots}) between our algorithm and the optimal algorithm when varying the number of new tasks. The error bars represent the min/max values of the simulations.} 
    \label{fig:vary_missions}
    \vspace{-1.75em}
\end{figure}

We compare our token-based allocation scheme with the optimal algorithm, which produces the task assignment with the minimum cost by checking ${n \choose m}$ different assignments. We show the cost of the behavior of the final assignment and the computation time of the two algorithms. 

We varied the number of robots from 1 to 20 with 10 fixed new sub-tasks (Fig. \ref{fig:vary_robots}). Similarly, we varied the number of new sub-tasks from 1 to 8 with 5 robots (Fig. \ref{fig:vary_missions}). For each of these scenarios, we ran 30 simulations, randomizing the robots' capabilities and current tasks each time. The simulations ran on a 2.5 GHz quad-core Intel Core i7 CPU.

In both cases, the optimal algorithm's computation time grows exponentially. The computation time of our task allocation algorithm grows much slower while also maintaining little to no sub-optimality in the final allocation results.
% \hkc{is there more discussion? anything interesting to say beyond this one paragraph?}
\section{Conclusion}

We present an approach for robots to automatically distribute a new task while still satisfying their current tasks. Each robot determines if it can satisfy both its current task and the new sub-tasks and resynthesizes its  behavior accordingly. We provide a  heuristic token-based task  distribution algorithm to determine the final task assignment for the new task. The algorithm is scalable and provides a near-optimal assignment that minimizes overall cost.

In future work, we will consider new tasks that are reactive, which will require robots to be able to adapt their behavior at runtime, and a method to model the dynamic and possibly adversarial environment. We also plan to introduce new tasks that require collaboration between robots, which will add complexity to both the synthesis of new behaviors and the allocation of tasks.

% incorporating safety constraints and a collision avoidance scheme for the robots, as well as considering new tasks that are reactive and require robots to collaborate to perform. 

% \include{rough_approach}

%%%%%%%%%%%%%%%%%%%%%%%

%%%%%%%%
\bibliography{references}
\bibliographystyle{ieeetr}
\end{document}